\documentclass[]{elsarticle}
\usepackage[colorlinks=false,linkcolor=black, citecolor=black, urlcolor=black]{hyperref}

\usepackage{lineno,hyperref}
\modulolinenumbers[5]

\usepackage{graphics}
\usepackage{booktabs}
\usepackage{graphicx}
\usepackage{multirow}
\usepackage{algorithm}

\usepackage{tabularx} 
\usepackage{booktabs} 
\usepackage{amsmath}
\usepackage{multirow}
\usepackage{listings}
\usepackage{color}
\usepackage{subfigure}
\usepackage{algorithm}
\usepackage{algpseudocode}

\journal{Arxiv}
\date{}

\begin{document}

\begin{frontmatter}

\title{DeepVS: An Efficient and Generic Approach for Source Code Modeling Usage}






\author[add1]{Yasir Hussain\corref{cor1}}
\ead{yaxirhuxxain@nuaa.edu.cn}
\author[add1,add2,add3]{Zhiqiu Huang\corref{cor1}}
\ead{zqhuang@nuaa.edu.cn}
\author[add1]{Yu Zhou}
\ead{zhouyu@nuaa.edu.cn}
\author[add1]{Senzhang Wang}
\ead{szwang@nuaa.edu.cn}

\address[add1]{College of Computer Science and Technology, Nanjing University of Aeronautics and Astronautics (NUAA), Nanjing 211106, China}
\address[add2]{Key Laboratory of Safety-Critical Software, NUAA, Ministry of Industry and Information Technology, Nanjing 211106, China}
\address[add3]{Collaborative Innovation Center of Novel Software Technology and Industrialization, Nanjing 210093, China}

\cortext[cor1]{Corresponding author}

\begin{abstract}
	The source code suggestions provided by current IDEs are mostly dependent on static type learning. These suggestions often end up proposing irrelevant suggestions for a peculiar context. Recently, deep learning-based approaches have shown great potential in the modeling of source code for various software engineering tasks. However, these techniques lack adequate generalization and resistance to acclimate the use of such models in a real-world software development environment. This letter presents \textit{DeepVS}, an end-to-end deep neural code completion tool that learns from existing codebases by exploiting the bidirectional Gated Recurrent Unit (BiGRU) neural net. The proposed tool is capable of providing source code suggestions instantly in an IDE by using pre-trained BiGRU neural net. The evaluation of this work is two-fold, quantitative and qualitative. Through extensive evaluation on ten real-world open-source software systems, the proposed method shows significant performance enhancement and its practicality. Moreover, the results also suggest that \textit{DeepVS} tool is capable of suggesting zero-day (unseen) code tokens by learning coding patterns from real-world software systems.
\end{abstract}

\end{frontmatter}

\section{Introduction} \label{Introduction}
Source code suggestion and completion are vital features of an integrated development environment (IDE). Software developers excessively rely on such features while developing software. Most of the modern IDE’s source code completion tools \cite{bruch2009learning} provide the next possible source code by leveraging the information already exists in the code editor of the IDE. Due to the limited historical data available in the code editor, the IDE may not be able to provide the correct suggestions or possibly torment the developer with irrelevant suggestions. 

Recently, deep learning techniques have been widely applied for various source code modeling tasks such as source code completion \cite{raychev2014code,white2015toward}, error fixing \cite{gupta2018deep,santos2018syntax}, method naming \cite{alon2019code2vec}, etc.  These approaches \cite{raychev2014code,white2015toward} have shown how they can significantly improve such tools by learning from historical codebases (e.g. GitHub). However, there are various limitations and concerns:
\begin{itemize}
	\item Most of these techniques lack to capture long-term context dependencies of source code resulting in subordinate performance.
	\item Most techniques are usually formed from recognized natural events and
	therefore, are incapable of suggesting zero-day (unseen) code tokens.
	\item Most methods require a huge amount of computation power as well as time to train such models. 
	\item Current techniques lack to illustrate, how they can assist software developers in a real-world software development environment.
	\item The existing works present no enlightenment, how these models can be integrated into an IDE.
\end{itemize}
To defy the above apprehensions, we propose a neural code completion tool named \textit{DeepVS}. The proposed tool leverages the bi-directional neural gating mechanism to learn coding patterns from existing software systems (codebase) resulting in a significant performance boost. This letter makes the subsequent contributions:

\begin{itemize}
	\item An end-to-end neural code completion tool trained on ten real-world open-source software systems with over 13 million code tokens. 
	
	\item The projected method is capable of providing suggestions in real-time directly in an IDE. Further, \textit{DeepVS} tool is capable of providing zero-day code tokens. The \textit{DeepVS} tool is publicly available at \url{https://github.com/yaxirhuxxain/DeepVS/}.
	
	\item The quantitative evaluation with ten real-world open-source software systems and qualitative analysis of \textit{DeepVS} tool verifies that the projected method is accurate and subordinate the existing approaches.
	
\end{itemize}

\section{Approach}
The projected method begins by preprocessing the codebase. First, we perform standardization to obtain the optimal feature set. To learn real-world coding patterns, we tokenize each source code file in the codebase into multiple sequences of fixed context. Next, for the purpose of neural language model training, we vectorize these extracted contexts. Finally, we train and test the BiGRU classifier for the code completion task.

\section{Problem Statement}
Given a source code file, the predictions for the next source code token could be defined as: $Y=f(\tau)$ where $Y$ represents the set of next source code suggestions, $f$ represents the classification function that predicts the next possible source code tokens and $\tau$ represents the context which is an input to the classification function. We use the source code of real-world software systems extracted from GitHub\footnote{https://www.github.com/ , accessed on 1/12/2019.} to build our codebase for the purpose of neural model training and testing. The collected codebase ($CB$) can be formalized as: $CB=<SS_1,SS_2,...,SS_i>$ and $SS_i=<cf_1,cf_2,...,cf_j>$, where $SS_1,SS_2,...,SS_i$ represents the $i$ software systems involved in $CB$ and $cf_1,cf_2,...,cf_j$ represents the $j$ code files involved in $i$th software systems.

\section{Preprocessing and Feature Extraction} \label{Preprocessing}
Preprocessing is a vital process in building neural language models. Following grassroots, we perform standardization, context extraction, and vectorization. 

\section{Standardization}
To standardize $CB$, we remove all blank lines, inline and block-level comments from each software system. Further, we replace all literal values to their generic data types to build the optimum feature set.

\section{Context Extraction}
To learn code patterns from historical software systems, each source code file in the codebase is parted into a sequence of space-separated tokens. These sequences are then subdivided into multiple sequences of fixed context ($\tau$).

\section{Vectorization}
To train the neural language model, we need to build a global vocabulary system. For this intend, we replace all singleton code tokens with a special token ("UNK"). Next, we build the vocabulary $V= <v_1,v_2,...,v_k>$ where $v_1,v_2,...,v_k$ represents $k$ unique source code token corresponds to an entry in the vocabulary. Then, each code token is replaced with its corresponding vocabulary index (positive integer) to convert context vectors into a form that is suitable for neural language model training. The vectorized source code can be expressed as: $SS_i^v= <\tau_1,\tau_2,...,\tau_k>$ where $\tau_1,\tau_2,...,\tau_k$ represents the $k$ number of context vectors found in $i$th software systems.

\section{Bi-Directional Context Learner}
We adopt GRU \cite{cho2014learning} which is a superior type of recurrent neural network (RNN) for the purpose of code completion classification. Conventional RNN suffers from fading gradient issue, limiting it from learning long-term context dependencies. Similar to LSTM, the GRU uses gating approach that exposes full hidden context on each time step $i$, which advances it to learn long-term context dependencies by an excessive edge. It is composed of two gates, the rest gate $r_i$ and the update gate $z_i$. It can be expressed as
\begin{equation}
h_i = (1-z_i)h_{i-1}+z_i\hbar_i
\end{equation}

Where $h$ and $\hbar$ is prior context and fresh context respectively.

\begin{equation}
z_i = \phi(W_z\tau_i+U_zh_{i-1})
\end{equation}

\begin{equation}
\hbar_i = tanh(W\tau_i+r_i \otimes Uh_{i-1})
\end{equation}

\begin{equation}
r_i = \phi(W_r\tau_i+U_rh_{i-1})
\end{equation}
The $\hbar$ is modulated by the reset gates $r_i$. Here $\otimes$ is element-wise multiplication and $\phi$ is the softmax activation function which can be expressed as 

\begin{equation}
S(Y) = \dfrac{e^Y}{\sum_{j}{e^{Yj}}}
\end{equation}

We utilize GRU in a bi-directional fashion to effectively learn the source code context bidirectionally. Further, we expect the neural learner to assign high probability to the correct source code suggestions by having low cross-entropy that can be expressed as

\begin{equation}
H_p(C) \approx -\dfrac{1}{N}\sum_{i=1}^{N}log_2 \text{\textit{ P}}(\tau^i|\tau^{i-1}_{i-n+1})
\end{equation}

\section{Tool Implementation}
\begin{figure}[H]
	\centering
	\includegraphics[width=\linewidth,keepaspectratio,clip,
	trim=0.6cm 5cm 0.6cm 2.5cm]{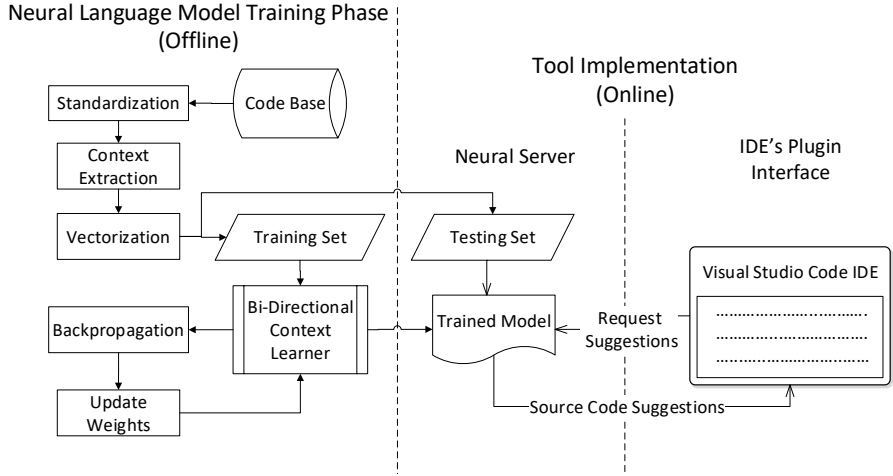}
	\caption{Overall workflow of DeepVS.}
	\label{fig:workflow}
\end{figure}

The trained model is saved, in order to use the model in our proposed tool. The main ingredient of this work is to serve the pre-trained BiGRU locally or over a cloud and use a plugin interface for real-time classification. Fig. \ref{fig:workflow} shows the overall workflow of the proposed approach where the offline part shows the steps involved in the pre-training BiGRU classifier and the online part illustrates the tool implementation details.

\section{Neural Server}
A cloud-based neural server is built to serve the pre-trained BiGRU neural language model as a centralized attendant to determine the likelihood scores for the next source code suggestions. In this work, the cloud platform \cite{armbrust2010view} is selected because of its centralized nature and scalability. The neural cloud server takes source code context $(\tau)$ as input and provides most likely next source code suggestions $(Y)$ by consulting the pre-trained BiGRU classifier.

\section{Deep Suggestion Engine}
As a proof of concept and demonstration purposes, we implement the plugin interface in \textit{Visual Studio Code} IDE. The \textit{DeepVS} tool can be triggered by using shortcut key \textit{CTRL+SPACE}. By triggering \textit{DeepVS}, it takes the source code prior to the current cursor position as context $(\tau)$ and requests \textit{Neural Server} for classification $Y=f(\tau)$. 

\section{Evaluation}
The evaluation of this work is two-fold, quantitative and qualitative. In quantitative evaluation, we assess the classification outcomes to test the performance of the proposed method and compare it with state-of-the-art methods. Metrics used to assess the quantitative efficiency of this work are top-k accuracy and mean reciprocal rank (MRR), which are mostly applied. For the qualitative evaluation of the work, we conduct an experimental study involving seven university students having software development experience. The experiment was performed with the help of a questionnaire. All the participants were demonstrated with the basic usage of the tool. The questionnaire given to the participants is shown in Table \ref{Table:Questionnaire}.

\begin{table}[tb]
	\scriptsize 
	\caption{Questionnaire used for evaluation.}
	\label{Table:Questionnaire}
	\begin{tabular}{|l|}
		\toprule
		\textbf{Questions}\\
		\midrule
		Q1: How easy it is to set up and use the DeepVS tool? \\
		Q2: Does \textit{DeepVS} tool provides suggestions in real-world coding environment?\\
		Q3: Does \textit{DeepVS} tool performs better than IDE's default suggestion engine?\\
		Q4: I will recommend \textit{DeepVS} to others.\\
		\bottomrule
	\end{tabular}
\end{table}

\section{Dataset} 
For the training and testing of the proposed method, we utilize the codebase anticipated in \cite{hindle2012naturalness,nguyen2018deep}. The codebase consists of ten open-source software systems (\textit{ant, cassandra, db40, jgit, poi, batik, antlr, itext, jts, maven}) containing over 13M code tokens with a large vocabulary of size 145,457 (singletons removed). Table \ref{Table:DataStatistics} shows the statistics of model training and testing sets in detail.

\begin{table}[htpb]
	\caption{Code database statistics}
	\label{Table:DataStatistics}
	\begin{tabular}{cccc}
		\toprule
		{}&{Line of Code (LoC)} & {Total Code Tokens} & {Vocab Size}\\
		\midrule
		Training & 1,684,158 & 12,086,347 &  \\
		Testing & 187,129 & 1,342,927 & \\
		\midrule
		Total & 1,871,287 & 13,429,274 & 145,457\\
		\bottomrule
	\end{tabular}
\end{table}

\section{Experimental Setup and Results}: 
The neural language model and cloud platform are build using \textit{Python 3.6}. The training of neural language models is done by using tensorflow4 v1.14 on Intel(R) Xeon(R) Silver 4110 CPU 2.10GHz with 32 cores and 128GB of ram running Ubuntu 18.04.2 LTS operating system, equipped with NVIDIA RTX 2080 GPU. The neural language model training is offline, thus has no impact on classification time. The pre-trained BiGRU model is used for classification purpose in \textit{neural server}. The \textit{Neural Serve} consists of a single core Intel CPU with 1GB Ram running Ubuntu-18 x64-bit non-GUI edition. The average classification time of the proposed approach is less than 20 milliseconds which is fairly adequate and acceptable.

\section{Quantitative Results}
\begin{figure}[htpb]
	\centering
	\includegraphics[width=0.8\linewidth]{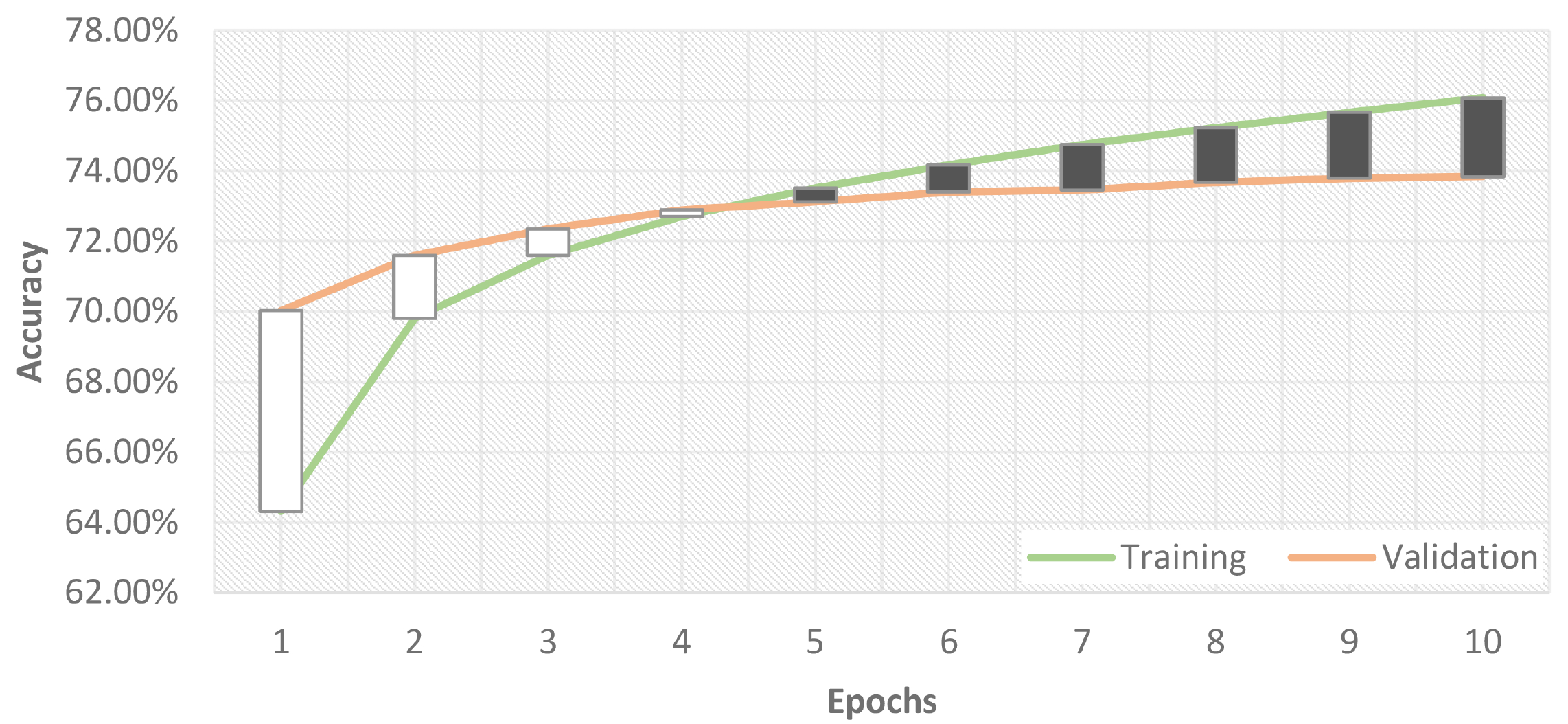}
	\caption{Training and validation accuracy comparison.}
	\label{fig:accuracy}
\end{figure}

\begin{figure}[htpb]
	\centering
	\includegraphics[width=0.8\linewidth]{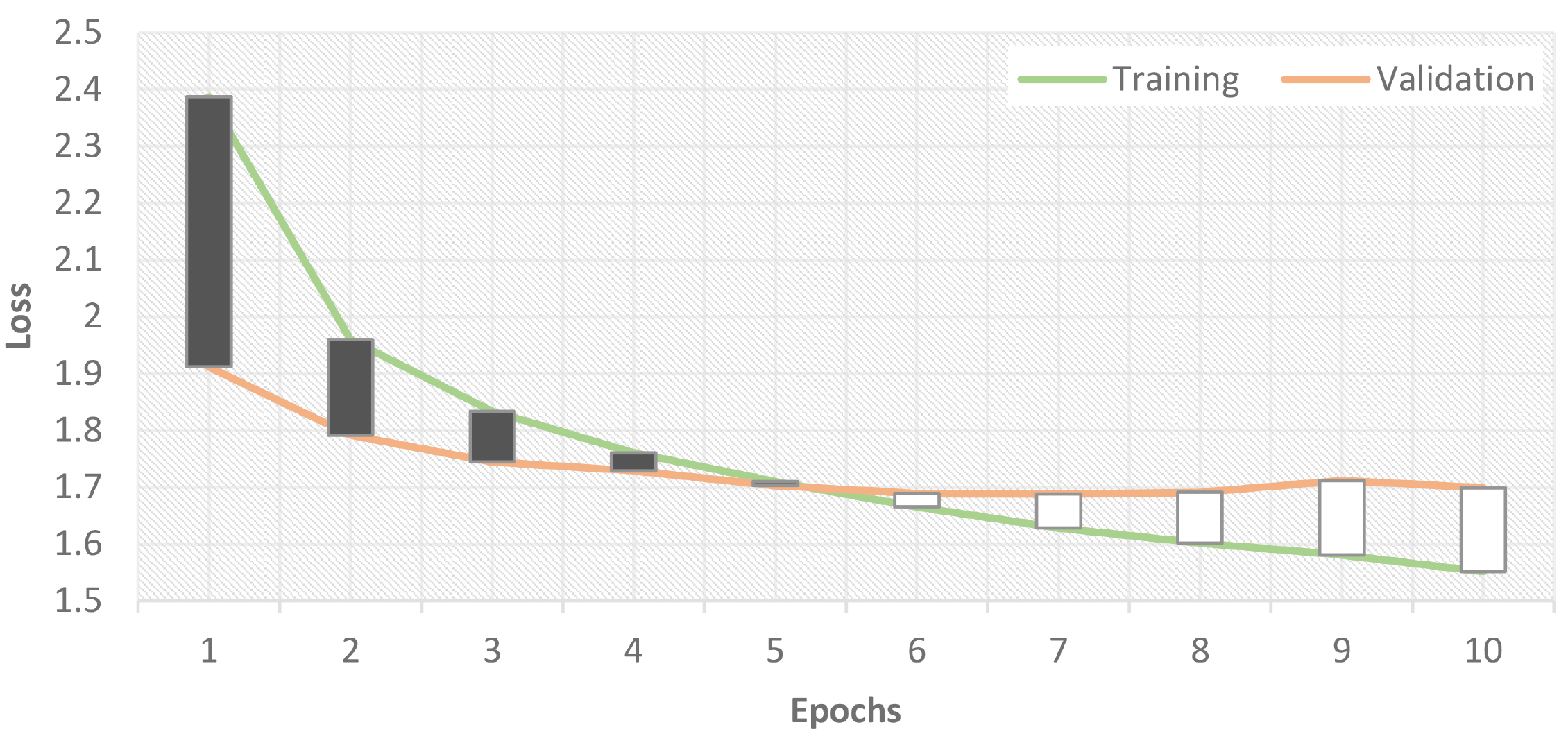}
	\caption{Training and validation loss comparison.}
	\label{fig:loss}
\end{figure}

In order to realistically evaluate the effectiveness of the proposed method, we compare the performance of our proposed method with other state-of-the-art methods. We choose Hindle el al. \cite{hindle2012naturalness} and White et al. \cite{white2015toward} work as baselines because they are related to our target task. The comprehensive results are given in Table \ref{Table:Results}. It can be witnessed that the proposed method proves noticeably better performance as compared to other baselines. We can observe that the proposed approach achieves 22.47\% higher accuracy as compared to the best baseline(RNN) by ranking the correct suggestions on the first index(Top-1). Fig. \ref{fig:accuracy} and Fig. \ref{fig:loss} represents the accuracy and loss of training and validation accordingly. 

\begin{table}[htpb]
	\caption{Accuracy and MRR scores comparison.}
	\label{Table:Results}
	\begin{tabular}{|c|c|c|c|c|c|}
		\toprule
		Model 	& \multicolumn{4}{c|}{Accuracy} & MRR \\
		\midrule
		& {Top-1} &  {Top-3} & {Top-5} & {Top-10} &  \\
		\midrule
		N-gram \cite{hindle2012naturalness}  & 48.47\% & 57.68\% & 59.87\% & 61.78\%  & 0.535 \\
		RNN \cite{white2015toward} & 51.30\%& 67.67\% & 72.18\% & 76.21\% &  0.596\\
		\textbf{BiGRU } & \textbf{73.77\%} & \textbf{84.39\%} & \textbf{86.97\%} & \textbf{89.34\%} & \textbf{0.792} \\
		\bottomrule
	\end{tabular}
\end{table}

\section{Qualitative Results}
\begin{figure}[htpb]
	\centering
	\includegraphics[width=0.7\linewidth]{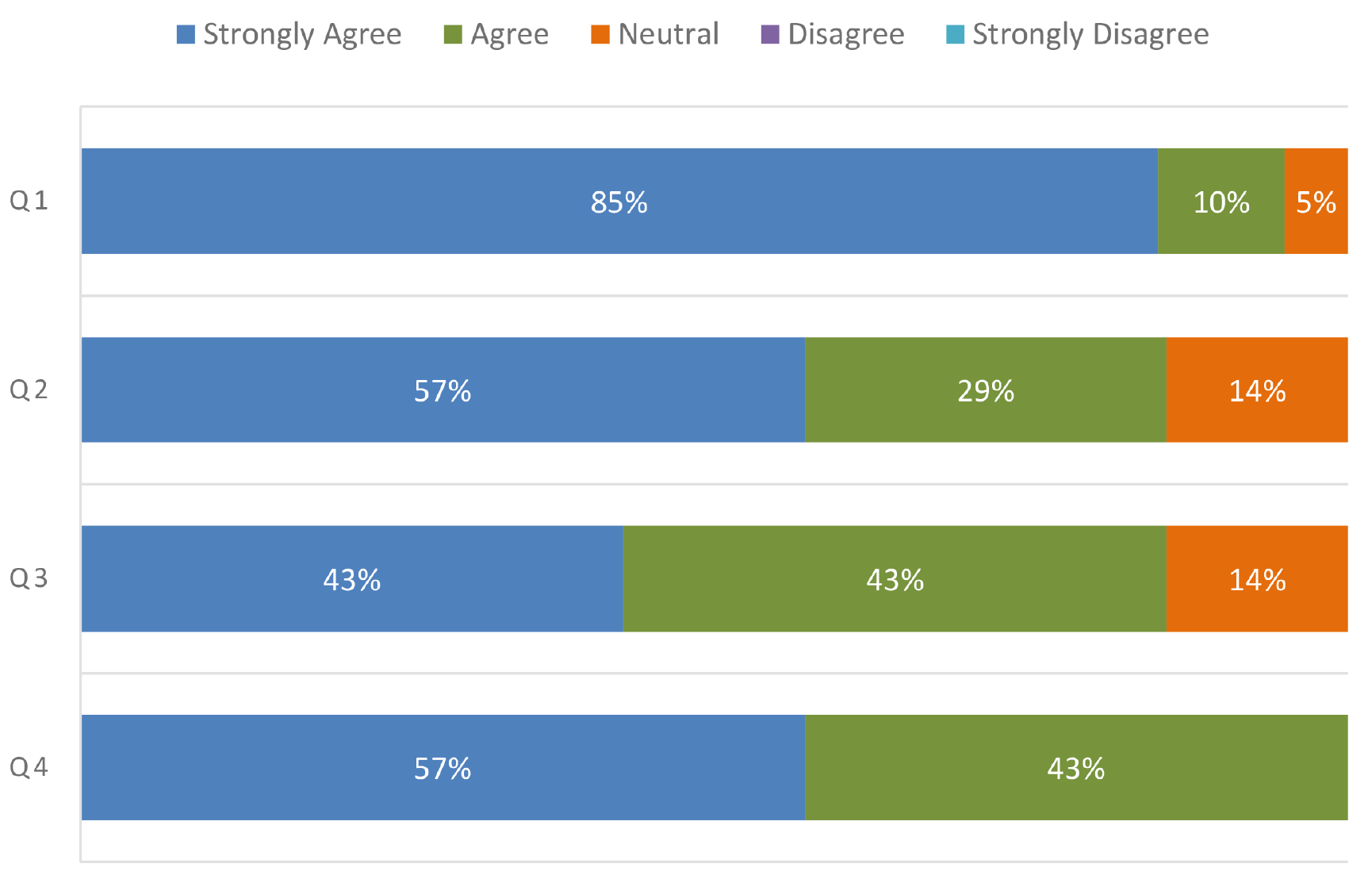}
	\caption{Questionnaire result.}
	\label{fig:Survey}
\end{figure}

As reported in Fig. \ref{fig:Survey}, participants found the \textit{DeepVS} tool is easy to set up and use. In Q2, all participants have agreed that \textit{DeepVS} tool is capable of providing suggestions in a real-world development environment. In Q3, out of seven participants three voted strongly agree, three of them voted agree and one voted neutral. In Q4, most of the participants have agreed that they would recommend \textit{DeepVS} to their peers. The illustrative examples of source code suggestion tasks are provided in Section \ref{DeepVSdemo}.

\section{DeepVS Tool Demo} \label{DeepVSdemo}
\begin{figure}[htbp]
	\centering
	\begin{subfigure}
		\centering
		\includegraphics[width=0.75\linewidth]{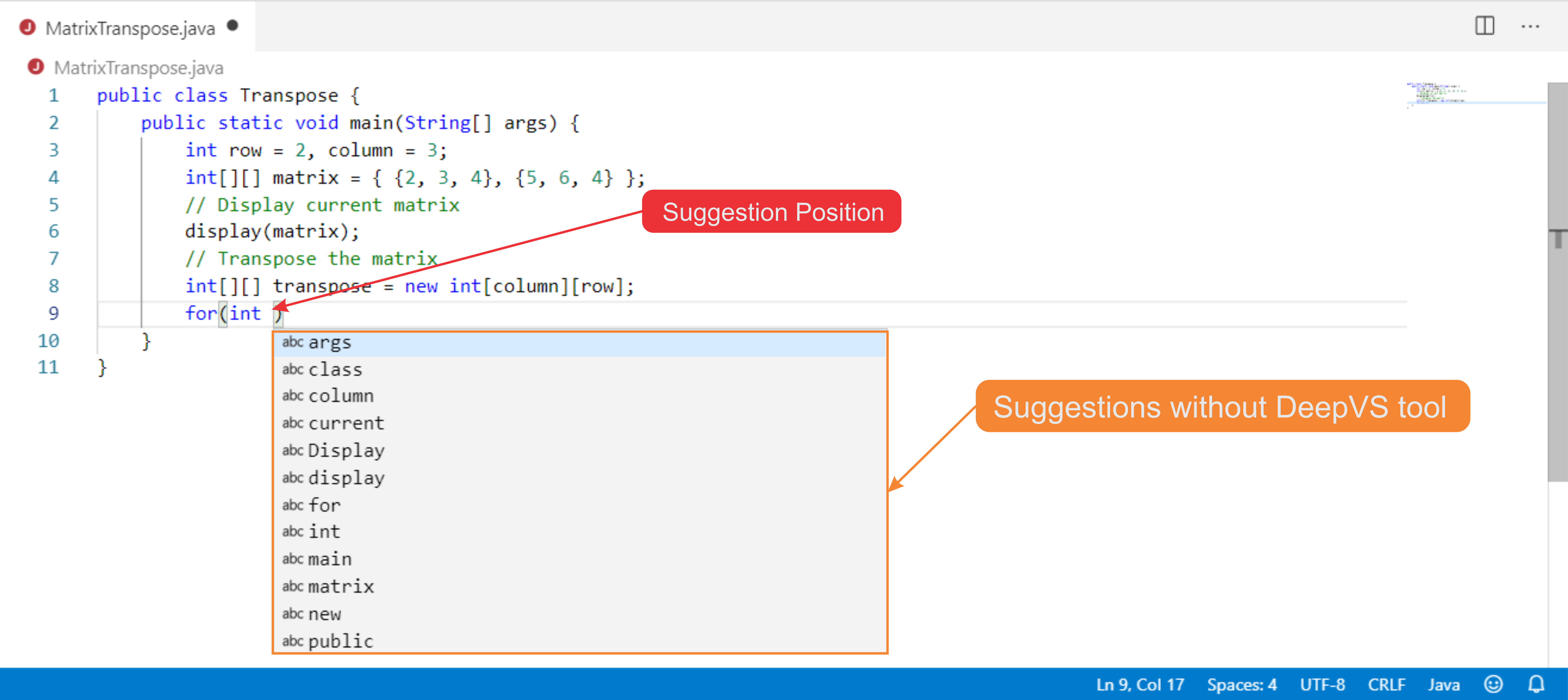}
		\label{fig:ex-a-1}
	\end{subfigure}%
	~ 
	\begin{subfigure}
		\centering
		\includegraphics[width=0.75\linewidth]{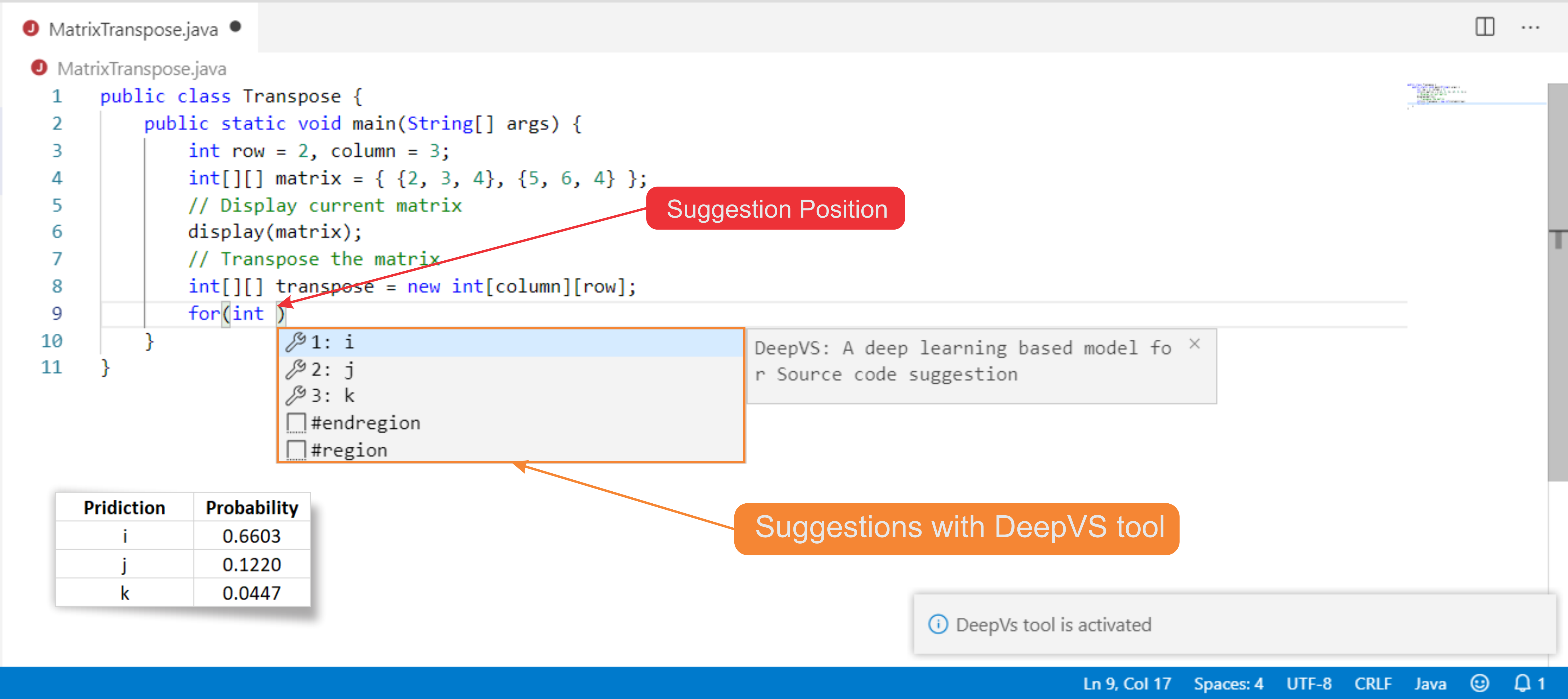}
		\label{fig:ex-a-2}
	\end{subfigure}
	\caption{Source code suggestion example with and without the \textit{DeepVs} tool along with their probabilities.}
	\label{fig:example-1}
\end{figure}

To show the effectiveness of \textit{DeepVS} tool for real-world usage, consider the example presented in Fig. \ref{fig:example-1} where a software developer is writing a complex program to find the transpose of a matrix. Hereby triggering (line \textit{9}) our \textit{DeepVS} tool it suggests the most probable next source code tokens \textit{i, j and k} where the \textit{Visual Studio Code's} default suggestion engine fails to provide any relevant suggestions instead, it torments the developer with irrelevant suggestions. Later on, at line \textit{21} the software developer is writing a print statement. Here the most probable next source code token is \textit{column} based on the given context. Hereby triggering our \textit{DeepVS} tool it suggests the correct next source code token \textit{column} at its first index whereas the IDE’s default source code suggestion tool ranks the correct suggestion on its third index. From the given examples, we can observe that the \textit{DeepVS} tool is capable of suggesting zero-day code tokens and have a better understanding of context while providing suggestions.

\begin{figure}[htbp]
	\centering
	\begin{subfigure}
		\centering
		\includegraphics[width=0.75\linewidth]{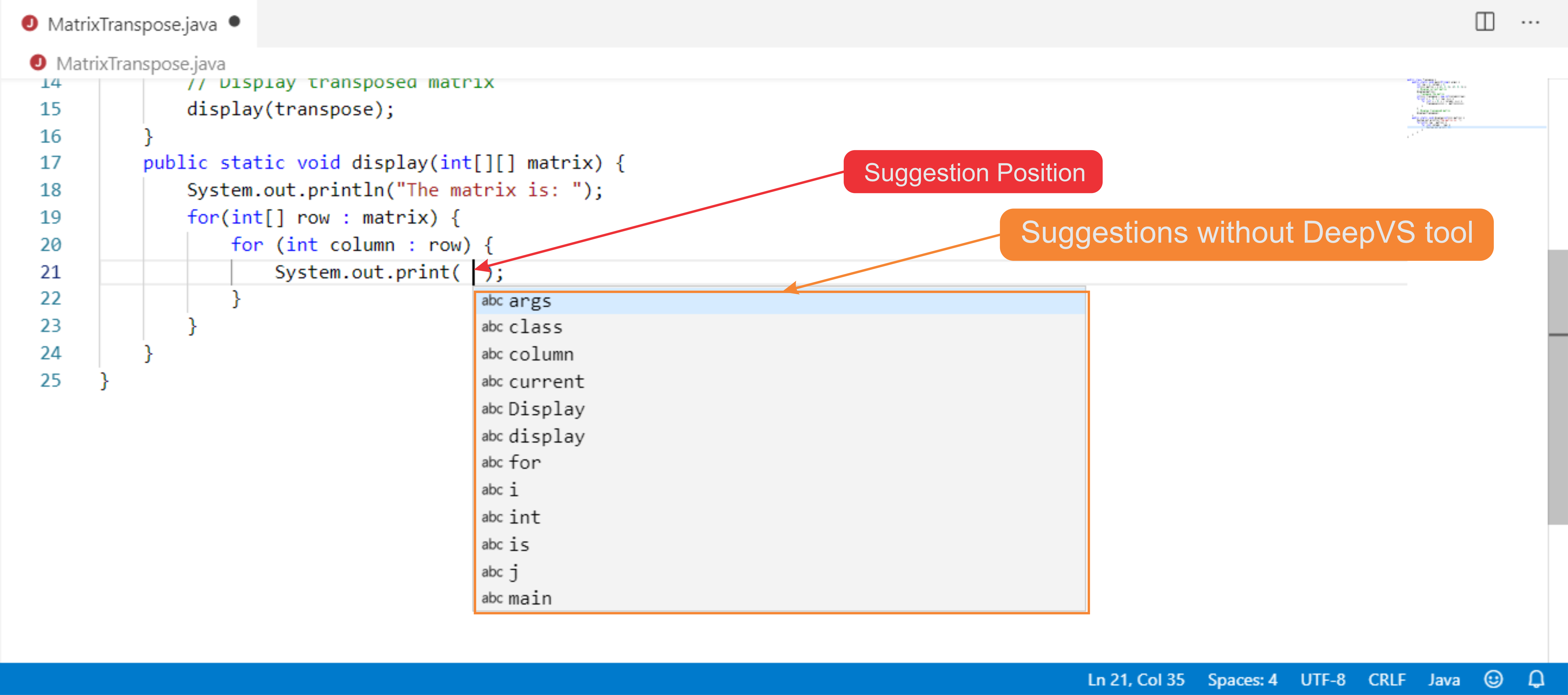}
		\label{fig:ex-b-1}
	\end{subfigure}%
	~ 
	\begin{subfigure}
		\centering
		\includegraphics[width=0.75\linewidth]{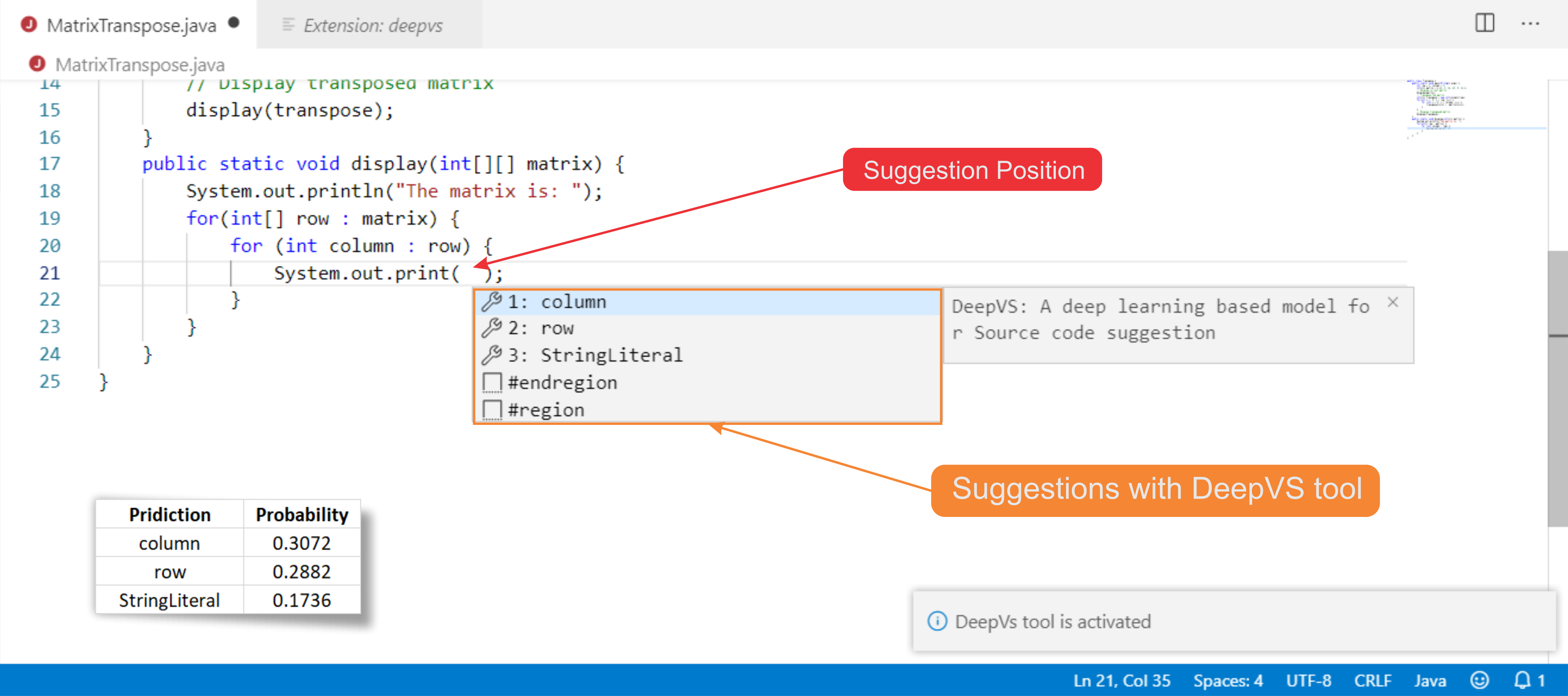}
		\label{fig:ex-b-2}
	\end{subfigure}
	\caption{Another Source code suggestion example along with their probabilities.}
	\label{fig:example-2}
\end{figure}

\section{Proposed Work's Major Benefits} 
The qualitative and quantitative evaluation verifies the feasibility and practicality of this work. We summarize the benefits of our proposed framework and the \textit{DeepVS} tool as follows:

\begin{itemize}
	
	\item To the best of our knowledge, the proposed approach is the first one that enables the usage of machine/deep learning-based source code models directly in an IDE.
	
	\item We present an end-to-end deep neural code completion tool named \textit{DeepVS} which verifies the practicality of the proposed approach.
	
	\item The \textit{DeepVS} tool is trained on ten real-world open-source software systems with over 13M code tokens, thus capable of suggesting zero-day (unseen) code tokens by leveraging large scale historical codebase.
	
	\item  Further, \textit{DeepVS} tool has better understanding of context while providing suggestions. The proposed tool ranks the suggestions effectively without tormenting the developer with unnecessary suggestion.
	
\end{itemize}

\section{Conclusion}
In this letter, we presented \textit{DeepVS}, an end-to-end deep neural code completion tool that leverages from the pre-trained BiGRU classifier to provide code suggestions directly in an IDE in real-time. The proposed tool is trained and tested on real-world software systems with over 13M code tokens. Further, we have demonstrated the effectiveness and practicality of the proposed tool in a real-world use case(IDE). Moreover, the approach illustrated in this work is general and can help enable the usage of neural language models for other source code modeling tasks directly in an IDE.

\section*{Acknowledgments}
This work was supported by the National Key R\&D (grant no. 2018YFB1003902), Natural Science Foundation of Jiangsu Province (No. BK20170809), National Natural Science Foundation of China (No. 61972197) and Qing Lan Project.

\bibliography{DeepVS}

\end{document}